\documentclass[sigconf]{acmart}
\AtBeginDocument{%
  \providecommand\BibTeX{{%
    \normalfont B\kern-0.5em{\scshape i\kern-0.25em b}\kern-0.8em\TeX}}}

\setcopyright{acmcopyright}
\copyrightyear{2021}
\acmYear{2021}
\acmDOI{}

\acmConference[]{}{}{}
\acmPrice{}
\acmISBN{}

\usepackage{algorithm}
\usepackage{algpseudocode}     

\begin{document}

\title{Efficient and Generic 1D Dilated Convolution Layer for Deep Learning}

\author{Narendra Chaudhary}
\orcid{https://orcid.org/0000-0002-0941-5945}
\affiliation{%
  \institution{Intel Labs}
  \city{Bangalore}
  \country{India}
  \postcode{560103}
}
\email{narendra.chaudhary@intel.com}

\author{Sanchit Misra}
\affiliation{%
  \institution{Intel Labs}
  \city{Bangalore}
  \country{India}}
\email{sanchit.misra@intel.com}

\author{Dhiraj Kalamkar}
\affiliation{%
  \institution{Intel Labs}
  \city{Bangalore}
  \country{India}
}
\email{dhiraj.d.kalamkar@intel.com}

\author{Alexander Heinecke}
\affiliation{%
 \institution{Intel Labs}
 \city{Santa Clara}
 \state{California}
 \country{USA}}
\email{alexander.heinecke@intel.com}

\author{Evangelos Georganas}
\affiliation{%
  \institution{Intel Labs}
  \city{Santa Clara}
 \state{California}
 \country{USA}}
\email{evangelos.georganas@intel.com}

\author{Barukh Ziv}
\affiliation{%
  \institution{Intel Corporation}
  \city{Haifa}
  \country{Israel}}
\email{barukh.ziv@intel.com}

\author{Menachem Adelman}
\affiliation{%
  \institution{Intel Corporation}
  \city{Haifa}
  \country{Israel}}
\email{menachem.adelman@intel.com}

\author{Bharat Kaul}
\affiliation{%
  \institution{Intel Labs}
  \city{Bangalore}
  \country{India}}
\email{bharat.kaul@intel.com}

\renewcommand{\shortauthors}{Chaudhary, Misra and Kalamkar, et al.}

\begin{abstract}
Convolutional neural networks (CNNs) have found many applications in tasks involving two-dimensional (2D) data, such as image classification and image processing. These networks use 2D convolution layers, and therefore, 2D convolution layers have been heavily optimized on CPUs and GPUs. However, in many applications -- for example genomics and speech recognition, the data can be one-dimensional (1D) or has one dimension significantly longer than the other dimensions. Such applications can benefit from optimized 1D convolution layers. In this work, we introduce our efficient implementation of a generic 1D convolution layer covering a wide range of input tensor widths, filter widths, number of channels, number of filters, and dilation parameters. It is optimized for x86 CPU architectures, in particular, for architectures containing Intel\textsuperscript{\textregistered} AVX-512 and AVX-512 BFloat16 instructions. We use the LIBXSMM library's batch-reduce General Matrix Multiplication (BRGEMM) kernel for single-precision (FP32) and brain floating-point (BFloat16) precision. We demonstrate that our implementation can achieve up to 80\% efficiency on Intel\textsuperscript{\textregistered} Xeon\textsuperscript{\textregistered} Cascade Lake and Cooper Lake  CPUs. Additionally, we show the generalization capability of our BRGEMM based approach by achieving high efficiency across a range of parameters. We consistently achieve higher efficiency than the 1D convolution layer with Intel\textsuperscript{\textregistered} oneDNN library backend for varying input tensor widths, filter widths, number of channels, filters, and dilation parameters. Finally, we demonstrate the performance of our optimized 1D convolution layer by utilizing it in the end-to-end neural network training with real genomics datasets and achieve up to $6.86\times$ speedup over the oneDNN library-based implementation on Cascade Lake CPUs. We also demonstrate the scaling with 16 sockets of Cascade/Cooper Lake CPUs and achieve significant speedup over eight V100 GPUs using a similar power envelop. In the end-to-end training, we get a speedup of $1.41\times$ on Cascade Lake with FP32, $1.57\times$ on Cooper Lake with FP32, and $2.27\times$ on Cooper Lake with BFloat16 over eight V100 GPUs with FP32. Our results demonstrate that software optimizations with Intel\textsuperscript{\textregistered} AVX-512, AVX-512 BFloat16 instructions can provide significant performance benefits and scale to deep learning applications.

\null
\noindent \textbf{Code Availability} - \url{https://github.com/hfp/libxsmm/tree/master/samples/deeplearning/conv1dopti_layer}
\end{abstract}





\maketitle

\section{Introduction}

Deep learning techniques can extract information from datasets of text, audio, images, and videos. These techniques have significantly improved performance in image classification \cite{krizhevsky2012imagenet, he2016deep, lecun2015deep}, image denoising \cite{zhang2017beyond,jin2017deep}, and natural language translation \cite{wu2016google} problems. Deep learning has become viable due to growth in dataset sizes and computing power. It is increasingly being employed in emerging fields with tremendous growth in dataset sizes, such as computational genomics \cite{poplin2018universal, rai2020single, Lal829481}. It is expected that the computing needs \cite{thompson2020computational, hernandez2020measuring} of deep learning algorithms will grow faster than Moore's law. At the same time, Leiserson et. al. \cite{Leisersoneaam9744} has argued that improvements in software, algorithm, and hardware architecture can provide higher than Moore's law \cite{mack2011fifty} speedup to applications. Therefore, deep learning algorithms need optimized implementations to keep pace with their computing demands. One of the key computational kernel in deep learning applications is the convolution layer. Deep convolutional neural networks (CNNs) consist of multiple compute-intensive convolution layers. CNNs with two-dimensional (2D) convolution layers have found applications in several image classification \cite{krizhevsky2012imagenet, he2016deep,simonyan2014very}, denoising \cite{zhang2017beyond,jin2017deep}, superresolution \cite{dong2015image,wang2020deep}, and segmentation \cite{minaee2021image} tasks. Therefore, GPU/CPU implementations of CNNs have been heavily optimized for two-dimensional (2D) image data. 


However, many datasets in nature are one-dimensional (1D) or have one dimension longer than the other dimensions. Audio, speech, text, and genomic sequencing datasets are some examples of 1D datasets. Deep learning applications involving 1D datasets frequently utilize 1D CNNs \cite{kiranyaz20211d}. 1D CNNs have been used for large scale audio classification \cite{gemmeke2017audio, kong2020panns} and speech processing \cite{li2019jasper, kriman2020quartznet} tasks. Many 1D datasets can also have relationships spanning across long distances along the width. In such cases, CNNs need 1D convolution layers with a wide receptive field. 1D convolution layers with dilation \cite{yu2015multi} satisfy this need. Dilated convolution layers have been used in Google's Wavenet \cite{oord2016wavenet, oord2018parallel} architectures for text to audio generation. Recently, deep 1D CNNs have also been utilized to perform denoising \cite{rai2020single, Lal829481} and peak detection in Assay for Transposase-Accessible Chromatin sequence (ATAC-seq) data \cite{buenrostro2015atac}. Rai et al., \cite{rai2020single} used a UNet \cite{ronneberger2015u} type CNN architecture with 1D convolutional layers to upscale ATAC-seq data. Lal et al. \cite{Lal829481} used a 1D Resnet \cite{he2016deep} CNN with dilated convolution layers for simultaneous denoising and peak calling from low-coverage or low-quality ATAC-seq data.

Researchers have made several efforts to optimize 2D convolution layer kernels. Such as convolution using the image-to-column transform \cite{vasudevan2017parallel, anderson2017low}, the fast Fourier transform (FFT) \cite{mathieu2013fast, vasilache2014fast, zlateski2018fft} method and the Winograd \cite{lavin2016fast} method. These implementations make assumptions such as a 2D tensor data input, short filter sizes, local connectivity, and narrow receptive fields. Specialized library implementations, such as oneDNN \cite{onednn} and cuDNN \cite{chetlur2014cudnn} have efficient implementations for 2D convolution layer. These implementations perform well for 1D convolution layers with short input tensor widths and short filter widths (1 to 3). In such cases, 1D computation is analogous to a 2D computation with tensors of short height and long width. However, oneDNN based implementations quickly become inefficient when we increase the 1D tensor width or filter width. Thus, several applications that need a long receptive field (audio, genomics, speech) do not perform efficiently.           

Our goal is to improve the computational efficiency of the 1D dilated convolution layer on CPUs. We want to develop computational kernels that are generic enough to achieve high efficiency across a wide range of  parameters for 1D convolution layer. Recently, Georganas et. al. \cite{georganas2019high} have shown that a single computational kernel of batch-reduce GEneral Matrix Multiply (BRGEMM) can implement many popular deep learning algorithms, including direct convolutions. Additionally, BRGEMM and small GEMMs based implementation can achieve high efficiency on CPUs. The LIBXSMM library \cite{heinecke2016libxsmm} provides efficient implementation of BRGEMM and small GEMMs in C language and provides support for both FP32 and Bfloat16 precision levels. It generates optimal assembly code with AVX-512 and AVX-512 BFloat16 SIMD instructions where applicable using Just-in-time (JIT) code generation that provides more instruction reduction than manually written intrinsics based code.

In this work, we rewrite the algorithm for 1D convolution in terms of BRGEMM and small GEMM kernels. We use the LIBXSMM library and cache blocking to develop a highly efficient implementation and then integrate the C++ code into the PyTorch framework. We use the LIBXSMM library to implement the 1D dilated convolution layer in single-precision (FP32) and BFloat16 precision. The combination of BRGEMM, small GEMMs, JIT-based code generation, and cache blocking along the tensor width dimension results in an implementation that performs well across a range of convolution layer parameters. To demonstrate this, we implement a one layer network for each precision type and perform experiments while varying parameters of filters, channels, input width, filter size, and dilation. In our experiments, the forward pass and the backward pass kernels display high efficiency across parameters. Specifically, our implementations achieve significant speedup over the oneDNN library for 1D tensors with long widths and filter sizes greater than or equal to 5.

We also demonstrate the effectiveness of the optimized 1D dilated convolution layer in end-to-end CNN training using real genomics (ATAC-seq) datasets. We integrate our 1D convolution layer in the training workflow presented in ~\cite{Lal829481} for using CNN with dilated convolution layers for denoising and peak calling from low-coverage or low-quality 1D ATAC-seq data. Our experiments show up to $6.86\times$ speedup over the oneDNN library for end-to-end training on Intel\textsuperscript{\textregistered} Cascade Lake CPUs. We also scale the experiments to multiple sockets and longer chunks of 1D data. Our scaling experiments demonstrate close to linear speedup while scaling from 1 to 16 CPU sockets of Intel Cascade/Cooper Lake CPUs. We compare the performance of 16 CPU sockets with that reported in ~\cite{Lal829481} for an Nvidia DGX-1 box (8 V100 GPUs with a host CPU)~\cite{nvidia-dgx1} since they are in the similar power envelop. In the end-to-end training with 16 CPU sockets, we get a speedup of $1.41\times$ on Cascade Lake with FP32, $1.57\times$ on Cooper Lake with FP32, and $2.27\times$ on Cooper Lake with BFloat16 over the DGX-1 box using FP32.

The paper is organized as follows. In Section ~\ref{sec:1d-conv}, we describe the 1D dilated convolution layer followed by our proposed approach for accelerating it including algorithms for forward and backward pass kernels in Section ~\ref{sec:our-work}. In Section ~\ref{sec:experiments}, we present the experiments on efficiency of the convolution layer, end-to-end CNN training, and scaling. In Section ~\ref{sec:concl}, we conclude the paper and mention ongoing work.

\section{1D Dilated Convolution Layer}
\label{sec:1d-conv}

One-dimensional (1D) convolution operation applies a 1D filter to a 1D input signal and produces a 1D output signal. However, in deep learning frameworks such as PyTorch and Tensorflow, a 1D convolution layer usually operates on a three-dimensional tensor input containing dimensions of batch size ($N$), input channels ($C$), and input width ($W$). Thus, input tensor ($In$) has a size of $(N, C, W)$. Similarly, if the convolution layer has $K$ number of filters, then the output tensor ($Out$) dimension becomes $(N, K, Q$) with output width as $Q$. We employ multithreading across the batch dimension ($N$) in the forward pass and the backward pass kernels. Thus, for further discussion, we will ignore the batch dimension and assume two-dimensional tensors for inputs and outputs. Hence, input tensor dimension changes to ($C, W$) and output tensor dimension changes to ($K,Q$). If the filter width is $S$ than the weight tensor ($Weight$) has size as $(K,C,S)$. We can represent the standard 1D convolution layer with parameters equation (\ref{eq:1d-conv}).
\begin{equation}
\label{eq:1d-conv}
    Out(k, q) = \sum_{c} \sum_{w + s = q} In(c,w) * Weight(k,c,s)
\end{equation}
In the 1D dilated convolution layer with a dilation amount of $d$, the filter weights are multiplied with every $dth$ element of the input tensor along the width dimension. We can represent the 1D dilated convolution layer by equation (\ref{eq:1d-dilated-conv}).
\begin{equation}
\label{eq:1d-dilated-conv}
    Out(k, q) = \sum_{c} \sum_{w + d*s = q} In(c,w) * Weight(k,c,s)
\end{equation}

\begin{figure}
  \centering
  \includegraphics[scale=0.5]{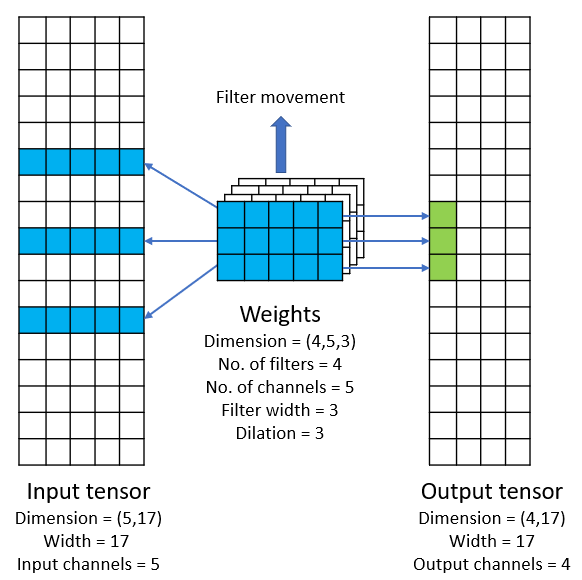}
  \caption{An example of the 1D dilated convolution layer with input channels ($C$) = 5, input width ($W$) = 17, number of filters ($K$) = 4, filter width ($S$) = 3, output width ($Q$) = 17, and dilation parameter ($d$) = 3.}
  \label{fig:1d dilated layer}
\end{figure}

Dilation in convolution increases the span of filters without increasing the number of weight parameters in them. Hence, dilated convolution increases the receptive field of a neural network without increasing its computational cost. We can also observe from equations (\ref{eq:1d-conv}) and (\ref{eq:1d-dilated-conv}) that the standard 1D convolution can be thought of as 1D dilated convolution with dilation parameter ($d$) equal to 1. Figure~\ref{fig:1d dilated layer} illustrates an example of the 1D dilated convolution layer with parameters of dilation, input width, input channels, and the number of filters. We assume zero-padding at the tensor edges.

\section{Our Efficient 1D Dilated Convolution Layer}
\label{sec:our-work}

We implement the forward pass and the backward data pass of the 1D dilated convolution layer using BRGEMM kernel of the LIBXSMM library. The backward weight pass kernel is implemented using small GEMM kernels. We do not implement the bias calculation of the forward and the backward pass but instead use the framework's implementation. BRGEMM kernel multiplies two matrix blocks $A_i \in \mathbb{R}^{m \times k}$ and $B_i \in \mathbb{R}^{k \times n}$ and reduces the partial results to a block $C_j \in \mathbb{R}^{m \times n}$ of a tensor C. The blocks $A_i$ and $B_i$ can be taken from any position in the larger A and B input tensors. BRGEMM kernel needs the following arguments: ($i$) Arrays of pointers for the $A_i$ and $B_i$ blocks to be multiplied, $(ii)$ a pointer to the output block $C_j$, $(iii)$ Size of blocks, $(iv)$ the number $(l_{br})$ of the blocks to be multiplied and $(v)$ the scaling parameters $\alpha$ and $\beta$. Equation (3) shows the batch reduce GEMM operation.
\begin{equation}
     C_j = \beta * C_j + \alpha * \sum_{0}^{l_{br} - 1} A_i * B_i
\end{equation}

Figure~2 illustrates batch-reduce GEMM operation with two-dimensional tensors. As shown in the figure, we can choose matrix blocks from any place in the tensor by specifying pointers and block sizes. It is also possible for the matrix blocks to overlap. In the following subsections, we present the forward pass, the backward data pass, and the backward weight pass algorithms with some figures for further explanation. 
\begin{figure}
  \centering
  \includegraphics[scale=0.37]{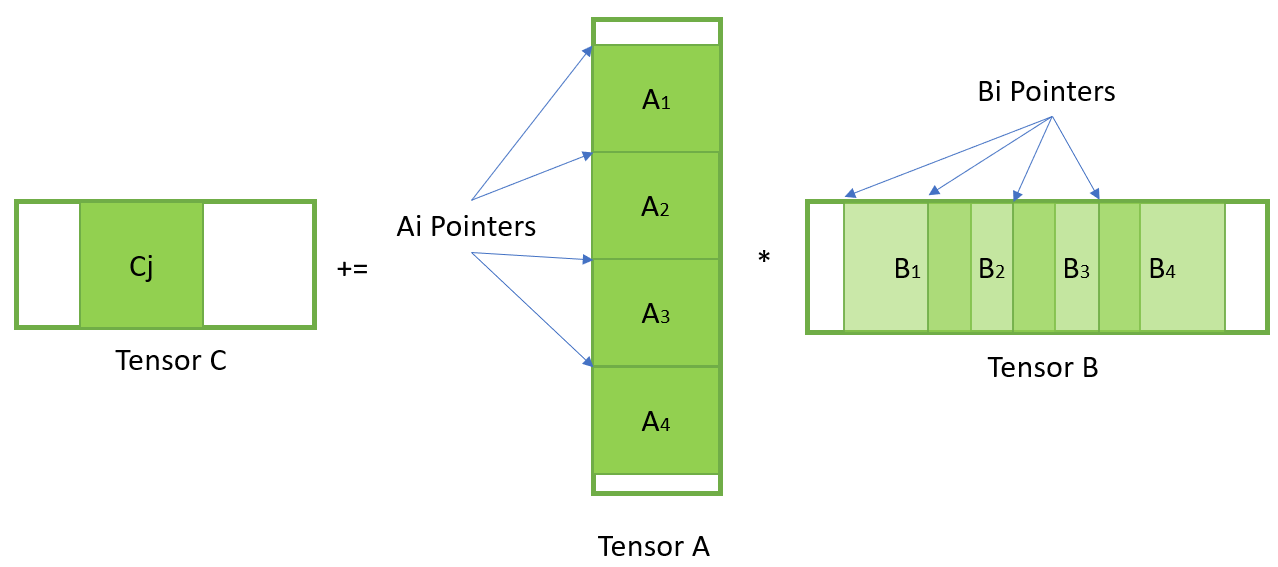}
  \caption{Example of Batch-reduce GEMM with 2D tensors. Blocks at $A_i$ and $B_i$ positions get multiplied and reduced to the $C_j$ block.}
  \label{fig:brGEMM}
\end{figure}

\subsection{Forward Pass}

\begin{figure*}[!tb]
  \centering
  \includegraphics[scale=0.6]{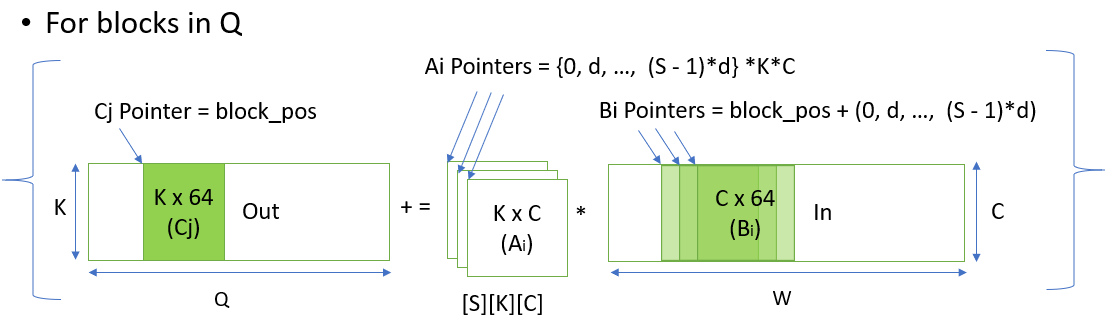}
  \caption{Forward pass kernel using batch-reduce GEMM. We multiply $A_i$ blocks from the weight tensor and $B_i$ blocks from the input tensor. The result is reduced into $C_j$ block in the output tensor. Cache blocking occurs along the width dimension.}
  \label{fig:Forward Pass}
\end{figure*}

To implement the forward pass, we first make some changes in the weight tensor layout to convert the forward pass computation into a matrix multiplication. We change the layout of the weight tensor from $(K, C, S)$ to $(S, K, C)$. Consequently, the 1D dilated convolution can be described by a series of $S$ GEMM operations explained in algorithm 1.
\begin{algorithm}
  \caption{Forward pass using GEMM operations}\label{Forward Pass GEMM}
  \begin{flushleft}
        \textbf{Inputs:} $In \in \mathbb{R}^{C \times W}$, $Weight \in \mathbb{R}^{S \times K \times C}$, $d \in \mathbb{R}$\\
        \textbf{Output:} $Out \in \mathbb{R}^{K \times Q}$
  \end{flushleft}
  \begin{algorithmic}[1]
    \Procedure{Forward pass}{$Out, In, Weight, d$} 
      \For{\texttt{$s = 0, 1, ..., S - 1$}}
        \State \texttt{$Out[:,:] \mathrel{{+}{=}} \mathbf{GEMM}(Weight[s, :, :], In[:, (d*s) : (d*s+Q)])$}
      \EndFor
      \State \textbf{return} $Out$\Comment{The output tensor}
    \EndProcedure
  \end{algorithmic}
\end{algorithm}

Once we can express the algorithm in terms of GEMM operations, we can convert it into BRGEMM operations. We can replace the for loop of algorithm~1 into a BRGEMM computation. However, a matrix problem-size suitable for the LIBXSMM library with $m,n,k$ matrix dimensions is $(m n k)^{1/3} <= 64$. The LIBXSMM library automatically employs an efficient utilization of the cache hierarchy when the $(m n k)^{1/3} <= 64$ condition is satisfied. Thus, we employ blocking along the input width dimension and perform BRGEMM operation on the blocks. In all our kernels, we keep the block length equal to 64 elements along the width dimension. Block length of 64 ensures that one dimension of the GEMM problem size remains within the LIBXSMM library's constraints. In our implementations, the other two dimensions are defined by the number of channels $C$ and the number of filters $K$ parameters. Thus, we achieve highly efficient cache optimized implementation whenever $(C*K)^{1/2} <= 64$. Additionally, the LIBXSMM library's GEMM kernels maintain good efficiency as long as the value of $(C*K)^{1/2}$ doesn't increase drastically. Algorithm~2 and figure~3 show the forward pass computation using BRGEMM kernel.

\begin{algorithm}
  \caption{Forward pass using BRGEMM operation}\label{Forward Pass BRGEMM}
  \begin{flushleft}
        \textbf{Inputs:} $In \in \mathbb{R}^{C \times W}$, $Weight \in \mathbb{R}^{S \times K \times C}$, $d \in \mathbb{R}$\\
        \textbf{Output:} $Out \in \mathbb{R}^{K \times Q}$
  \end{flushleft}
  \begin{algorithmic}[1]
    \Procedure{Forward pass}{$Out, In, Weight, d$} 
      \For{\texttt{$pos = 0 \ to \ Q \ in \ steps \ of \ 64$}}\Comment{Cache blocking}
        \For{$s = 0, 1, ..., S - 1$}\Comment{Generate pointers}
            \State \texttt{$A_{ptrs}[s] = Pointer \ to \ Weight[s, 0, 0]$}
            \State \texttt{$B_{ptrs}[s] = Pointer \ to \ In[0, (pos + s*d)]$}
        \EndFor
        \State \texttt{$\mathbf{BRGEMM}(A_{ptrs}, B_{ptrs}, Pointer \ to \ Out[0,pos], S)$}
      \EndFor
      \State \textbf{return} $Out$\Comment{The output tensor}
    \EndProcedure
  \end{algorithmic}
\end{algorithm}

\subsection{Backward Data Pass}

In the backward data pass implementation, we first change the weight tensor layout from $(K, C, S)$ to $(S, C, K)$. Similar to the forward pass, data gradient ($Grad_d \in \mathbb{R}^{C \times W}$) computation can be converted into a matrix multiplication of weights and output gradient ($Grad_{out} \in \mathbb{R}^{K \times Q}$). The backward data pass algorithm is very similar to the forward pass. We again employ the cache blocking along the width dimension with a block size of 64. We zero pad the gradient output ($Grad_{out}$) wherever needed. Algorithm~3 implements the backward data kernel using BRGEMM operation.

\begin{algorithm}
  \caption{Backward data pass using BRGEMM operation}\label{Backward Data Pass}
  \begin{flushleft}
        \textbf{Inputs:} $Grad_{out} \in \mathbb{R}^{K \times Q}$, $Weight \in \mathbb{R}^{S \times C \times K}$, $d \in \mathbb{R}$\\
        \textbf{Output:} $Grad_d \in \mathbb{R}^{C \times W}$
  \end{flushleft}
  \begin{algorithmic}[1]
    \Procedure{Backward data pass}{$Grad_d, Grad_{out}, Weight, d$} 
      \For{\texttt{$pos = 0 \ to \ W \ in \ steps \ of \ 64$}}\Comment{Cache blocking}
        \For{$s = 0, 1, ..., S - 1$}\Comment{Generate pointers}
            \State \texttt{$A_{ptrs}[s] = Pointer \ to \ Weight[s, 0, 0]$}
            \State \texttt{$B_{ptrs}[s] = Pointer \ to \ Grad_{out}[0, pos - (S - 1 - s)*d]$}
        \EndFor
        \State \texttt{$\mathbf{BRGEMM}(A_{ptrs}, B_{ptrs}, Pointer \ to \ Grad_d[0,pos], S)$}
      \EndFor
      \State \textbf{return} $Grad_d$\Comment{Data gradient}
    \EndProcedure
  \end{algorithmic}
\end{algorithm}

\subsection{Backward Weight Pass}

We utilize small GEMM operations in the backward weight pass implementation. We again do cache blocking along the width dimension with a block size of 64. The backward weight pass kernel can be less efficient than the other kernels because the data blocks cannot be kept in the cache for a long time. Additionally, the weight tensor must be shared across multiple threads when multithreading on the batch dimension $(N)$. Algorithm~4 implements the backward weight pass using small GEMM operations.

\begin{algorithm}
  \caption{Backward weight pass using  small GEMM operations}\label{Backward Weight Pass}
  \begin{flushleft}
        \textbf{Inputs:} $Grad_{out} \in \mathbb{R}^{K \times Q}$, $In \in \mathbb{R}^{C \times W}$, $d \in \mathbb{R}$\\
        \textbf{Output:} $Grad_w \in \mathbb{R}^{S \times C \times K}$
  \end{flushleft}
  \begin{algorithmic}[1]
    \Procedure{Backward weight pass}{$Grad_w, Grad_{out}, In, d$} 
      \For{\texttt{$pos = 0 \ to \ Q \ in \ steps \ of \ 64$}}\Comment{Cache blocking}
        \For{$s = 0, 1, ..., S - 1$}
            \State \texttt{$Grad_w[s,:,:] \mathrel{{+}{=}} \mathbf{GEMM}(In[:, (pos + s*d):(pos+s*d+64)], transpose(Grad_{out}[:, pos : (pos + 64)]))$}
        \EndFor
      \EndFor
      \State \textbf{return} $Grad_w$\Comment{Weight gradient}
    \EndProcedure
  \end{algorithmic}
\end{algorithm}

\section{Experiments and Results}
\label{sec:experiments}

In this section, we present details of our experiments and results. We use Intel\textsuperscript{\textregistered} Xeon\textsuperscript{\textregistered}~ Cascade Lake and Cooper Lake CPUs for all our experiments. Our first set of experiments focus on the efficiency and generality of the 1D convolution layer. We present the computational efficiency compared to peak machine performance of the forward pass and the backward pass implementations. We compare the efficiency of our single-precision and BFloat16 precision implementations with the oneDNN library. In our second experiment set, we conduct end-to-end training of a 1D Resnet CNN with 1D ATAC-seq data. Finally, we scale our experiments by increasing the number of compute sockets, dataset size, and ATAC-seq signal track size. We show multinode scaling results and compare them with multi-GPU results published in ~\cite{Lal829481}.                

\subsection{System Details}

In our Cascade Lake (CLX) experiments, we use Intel\textsuperscript{\textregistered} Xeon\textsuperscript{\textregistered} Platinum 8280 CPU @ 2.7 GHz. One socket of this CPU has 28 cores. It has L2 caches of 1 MegaByte (MB) per core and has an L3 cache of 38.5 MB. This CPU has a base frequency of 2.7 GHz and a single-core maximum turbo frequency of 4 GHz. We enable turbo for all cores during our experiments. This CPU supports AVX-512 instruction for single-precision computation with a peak machine performance of 4.3 TeraFLOPS.

In our Cooper Lake (CPX) experiments, we use Intel\textsuperscript{\textregistered} Xeon\textsuperscript{\textregistered} Platinum 8380HL CPU @ 2.9 GHz. One socket of this CPU has 28 cores. It has L2 caches of 1 MegaByte (MB) per core and has an L3 cache of 38.5 MB. This CPU has a base frequency of 2.9 GHz and a single-core maximum turbo frequency of 4.3 GHz. We enable turbo for all cores during our experiments. This CPU supports AVX-512 for single-precision and AVX-512 BFloat16 for Bfloat16 computation. This CPU has a peak machine performance of 4.66 TeraFLOPS for FP32 and 9.32 TeraFLOPS for BFloat16 computations.

\subsection{Experimental Details}
\label{sec:expt-details}

We use synthetic datasets for experiments to measure the efficiency of our optimized 1D convolution layer.

For end-to-end training experiments, we train a 1D Resnet CNN named AtacWorks \cite{Lal829481} and collect end-to-end training results with a genomics (ATAC-Seq) dataset. A trained AtacWorks neural network model takes noisy 1D ATAC-seq signal track segment as input and produces a corresponding 1D denoised signal track segment along with a 1D binary array of called peaks. Multiple loss functions are used to train the AtacWorks network. The loss function uses mean squared error (MSE) for the denoised signal and binary cross-entropy for the peak detection. The AtacWorks neural network architecture consists of multiple residual blocks, and each residual block contains 1D dilated convolution layers followed by a ReLU non-linearity. In total, AtacWorks utilizes 25 1D convolution layers to denoise and call peaks from the ATAC-seq signal. Most convolution layers have 15 channels, 15 filters, a filter size of 51, and a dilation of 8. AtacWorks network attempts to solve an important problem in genomics, and it is ideal for testing the computation performance of the 1D dilated convolution layer. AtacWorks is implemented for running 1D convolution on GPGPUs. For our experiments on CPUs, we integrate 1D convolution layers based on oneDNN and our optimized implementation into AtacWorks.

The training conditions are the same as that of ~\cite{Lal829481}, and the publicly available AtacWorks source code. We train the AtacWorks neural network for 25 epochs. We use chromosome 20 of the ATAC-seq dataset for validation and hold out chromosome 10 of the ATAC-seq dataset for testing. We use all other autosomes for training. The training set contains 32000 1D ATAC-seq signal track segments. Each signal track segment has a width of 50000, and it is padded on both sides by 5000 points to make the width of 60000. The validation set has 1280 signal track segments. 

\subsection{Efficiency Experiments}

We implement the forward and backward pass kernel using C++ language using the LIBXSMM library's functions and integrate the kernels into the PyTorch framework by writing a C++ extension. Subsequently, we can run the 1D dilated convolution layer in the PyTorch framework using our extension or oneDNN as backend. From the 1D convolution layer, we create a single layer convolutional neural network $(net)$, and time the forward and the backward pass for an average of 20 iterations. We provide the padded input tensor $(In \in \mathbb{R}^{N \times C \times W})$ to the forward pass, and sum the output tensor ($Out \in \mathbb{R}^{N \times K \times Q}$). Specifically, we time the \textit{Out = net.forward(In)} method for the forward pass, and the \textit{Out.sum().backward()} method for the backward pass. 
The computation times have some framework overhead. However, we still use this workflow to assess approximate performance comparison of different implementations of the 1D dilated convolution layer. We collect computation efficiency results on a single socket of Cascade Lake and Cooper Lake CPUs. 
The optimized C++ backend performs multi-threading along the batch dimension, therefore, giving the best performance when batch size is integer multiple of core count. The oneDNN backend provides the best performance when batch size is a power of two.  So, we keep a batch size $(N)$ of 56 with the optimized C++ backend and a batch size of 64 with the oneDNN backend.  

\begin{figure*}
  \centering
  \includegraphics[scale=0.52]{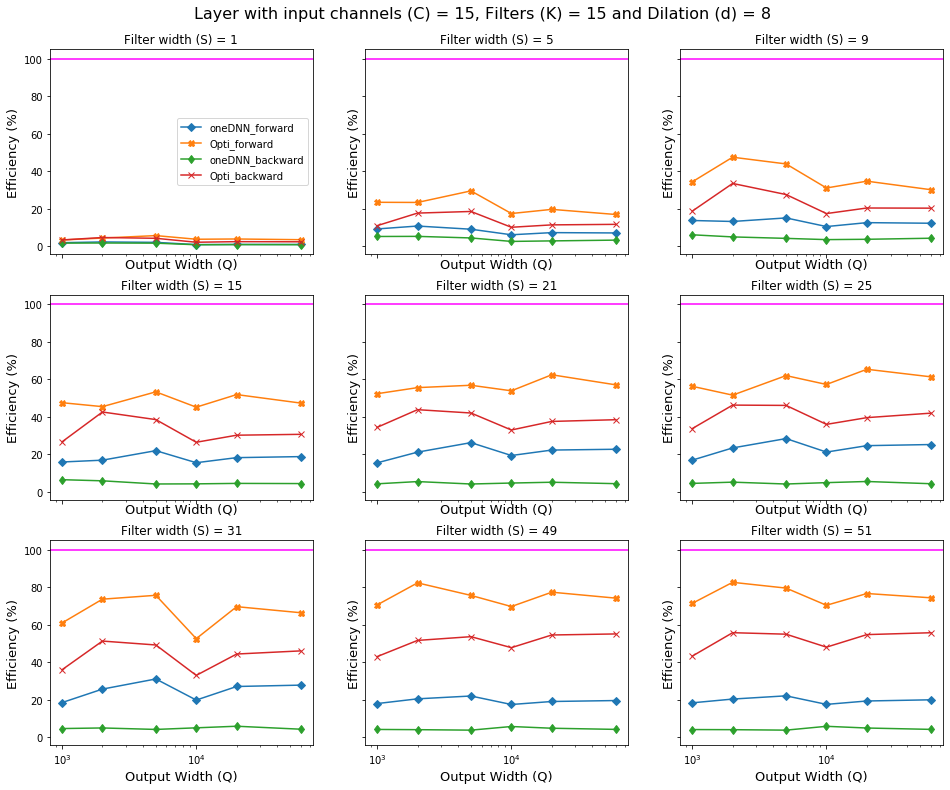}
  \caption{Plots for 1D convolution layer using FP32 with input channels (C) = 15, number of filters (K) = 15, and dilation (d) of 8 on single socket Cascade Lake. The efficiency of the forward pass and the backward pass computation is plotted with respect to output tensor width. }
  \label{fig:Efficiency_C15_F15_d8}
\end{figure*}

To check the efficiency and generality, we run the 1D dilated convolution layer for a wide range of parameters. We vary the output tensor width, number of channels, number of filters, filter widths, and the dilation parameter values. We choose output tensor width from the set \{1000, 2000, 5000, 10000, 20000, 60000\}, number of channels from the set \{1, 4, 8, 10, 15, 16, 32, 64\}, number of filters from the set \{1, 4, 8, 10, 15, 16, 32, 64\}, filter size from the set \{1, 5, 9, 15, 21, 25, 31, 49, 51\} and the dilation parameter from the set \{1, 2, 4, 8, 16\}. We use the output tensor width instead of the input tensor width because it remains constant for different filter sizes and dilation parameter values. 

Our experiments show that in most cases, optimized forward pass and backward pass computations achieve significantly higher efficiency compared to the oneDNN implementations. Specifically, our kernels are more efficient whenever the following condition is satisfied for the parameters of filter size (S), output tensor width (Q), number of channels (C), and number of filters (K).
\begin{equation}
\label{eq:condition}
    Condition \rightarrow (S >= 5) \land (Q >= 1000) \land (C > 1) \land (K > 1)
\end{equation}

\begin{figure*}
  \centering
  \includegraphics[scale=0.52]{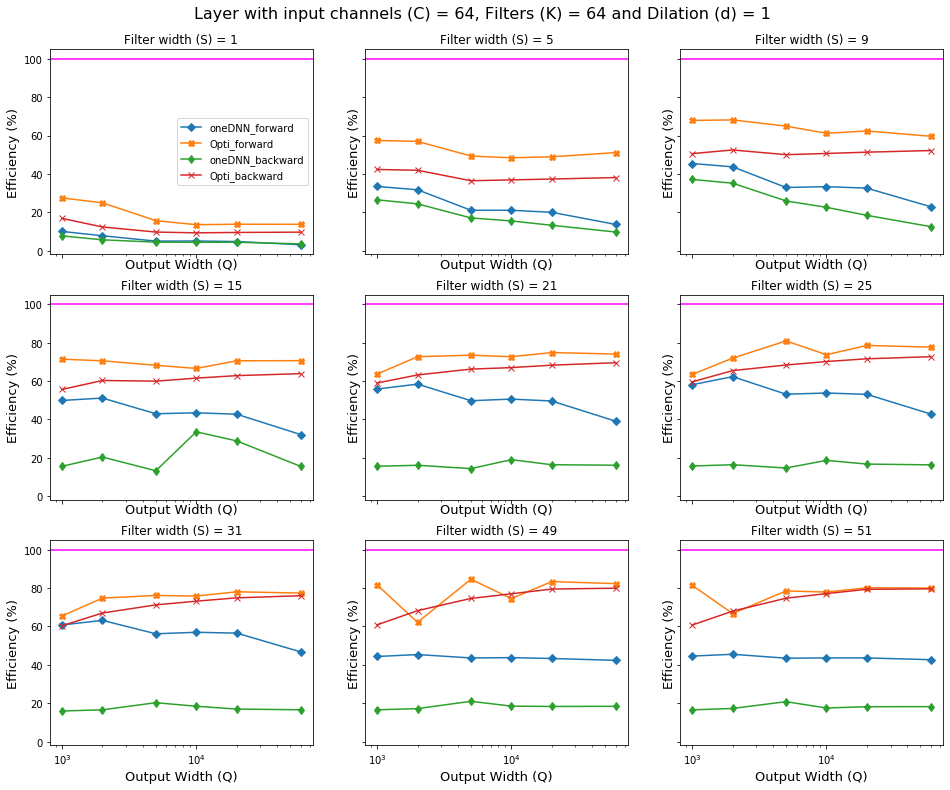}
  \caption{Plots for 1D convolution layer using FP32 with input channels (C) = 64, number of filters (K) = 64, and dilation (d) of 1 on single socket Cascade Lake. The efficiency of the forward pass and the backward pass computation is plotted with respect to output tensor width.}
  \label{fig:Efficiency_C64_F64_d1}
\end{figure*}

The optimization condition in equation~\ref{eq:condition} covers a wide range of parameters, and it shows that the optimized convolution layer is generic. It also demonstrates the effectiveness of BRGEMM, JIT code generation, and cache blocking along the width dimension. For the sake of brevity, we present here a few results and plots. Figure~\ref{fig:Efficiency_C64_F64_d1} plots show computational efficiency of the 1D dilated convolution layer with respect to the output tensor width. 
Plots in figure~\ref{fig:Efficiency_C15_F15_d8} are for the 1D dilated convolution layer with 15 input channels, 15 filters, and with the dilation parameter equal to 8. Each subplot in figure~\ref{fig:Efficiency_C15_F15_d8} corresponds to a different filter width. These results are obtained on a single-socket of Cascade Lake CPU for $20$ iterations. Figure~\ref{fig:Efficiency_C64_F64_d1} contains the efficiency results for a standard 1D convolution (dilation=1) with 64 channels and 64 filters. We observe that for larger filter widths, the forward pass and the backward pass can achieve up to 80\% efficiency. The optimized layer has the highest efficiency with larger filter widths and output tensor widths. Contrarily, the oneDNN based layer has less computation efficiency in those cases.   

\begin{figure*}
  \centering
  \includegraphics[scale=0.52]{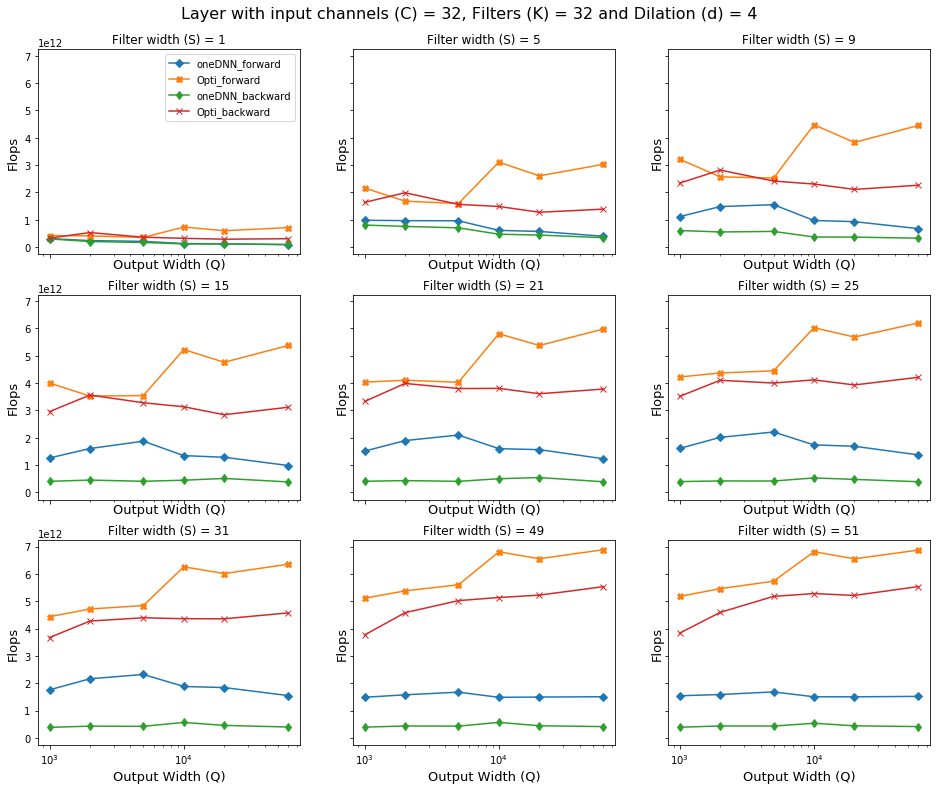}
  \caption{Plots for 1D convolution layer using BFloat16 with input channels (C) = 32, number of filters (K) = 32, and dilation (d) of 4 on single socket Cooper Lake. The performance (FLOPS) of the forward pass and the backward pass computation is plotted with respect to output tensor width.}
  \label{fig:Flops_C32_F32_d4}
\end{figure*}

We conduct similar experiments for the BFloat16 precision on a single-socket Cooper Lake machine. Figure~\ref{fig:Flops_C32_F32_d4} shows the plot of performance (FLOPS) with respect to output tensor width. In this experiment, convolution layers had 32 channels, 32 filters, and a dilation parameter of 4. In these plots, the oneDNN layer runs in single-precision, while our optimized convolution layer runs in BFloat16 precision. We can observe that BFloat16 implementation increases the performance in most cases. We get a $1.6\times$ speedup compared to the FP32 code. We again see the maximum performance with long output tensor widths and filter sizes. Our current implementation of the convolution layer in BFloat16 precision requires the input tensor width, the number of channels, and the number of filters to be even numbers. 

\subsection{End-to-End Training Experiment on a Single Socket CPU}
\label{}

In this experiment, we use the setup described in Section ~\ref{sec:expt-details} and train the AtacWorks neural network on a single socket of Cascade Lake or Cooper Lake CPUs. We reserve one CPU core on each socket for the PyTorch \textit{DataLoader()} worker and the remaining 27 CPU cores for computation. An attempt to utilize all 28 CPU cores on a socket for computation decreases efficiency. In such a case, the thread of \textit{DataLoader()} worker can slow down compute CPU cores by idling them. We use a batch size of 54 for the backend consisting of optimized code and a batch size of 64 with the oneDNN based backend. We use these batch sizes to obtain maximum training efficiency for each code. 

\begin{figure}
  \centering
  \includegraphics[scale=0.62]{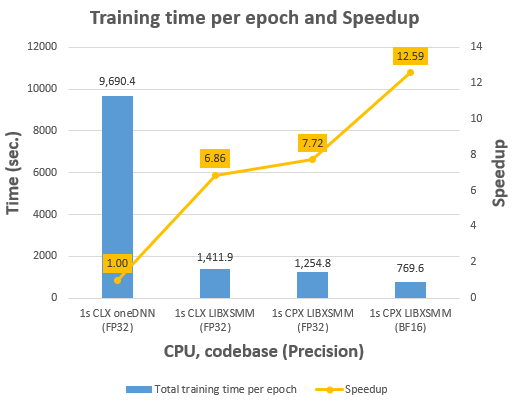}
  \caption{ATAC-seq training experiment results. Training time per epoch is an average of 25 epoch values during training. Speedup is over the baseline of oneDNN run on a single-socket Cascade Lake (CLX). Timing and speedup results are on a single-socket (28 cores) of Cascade Lake (CLX) and Cooper Lake (CPX).}
  \label{fig:single socket}
\end{figure}

Table~\ref{tab:single socket training} and figure~\ref{fig:single socket} show the end-to-end training time per epoch results on single-socket Cascade Lake (CLX) and Cooper Lake (CPX). To verify accuracy, we also train the network using original AtacWorks running on a V100 GPU with a batch size of 64.
We can observe that the implementation based on our optimized layer on a single-socket Cascade Lake achieves up to $6.86\times$ speedup over the oneDNN. Cooper Lake CPUs are slightly faster than Cascade Lake CPUs due to higher frequency and memory bandwidth. Additionally, we can observe from Table~\ref{tab:single socket training} that the training in BFloat16 precision can provide significant speedup without compromising the accuracy. In BFloat16 training experiments, most convolution layers of AtacWorks had 16 channels, 16 filters, a filter size of 51, and a dilation of 8. We also implemented a BFloat16 precision rectified linear unit (ReLU) layer using the LIBXSMM library to reduce time-consuming data conversion operations.

\begin{table}[hb]
  \caption{ATAC-seq training experiment results for 25 epochs. Training time per epoch is an average of 25 epoch values during training. Timing and accuracy results are on a single-socket (28 cores) of Cascade Lake (CLX) and Cooper Lake (CPX). Accuracy of original AtacWorks running on a single Nvidia V100 GPU is reported for comparison.}
  \label{tab:single socket training}
  \begin{tabular}{lccc}
    \toprule
    Device & Code & Training  & Accuracy \\
     & (Precision) & time (sec.) & (AUROC) \\
    \midrule
    1 V100 & CUDA (FP32) & \_ & 0.9386 \\
    1s CLX & oneDNN (FP32) & 9690.4 
    & 0.9388 \\
    1s CLX & LIBXSMM (FP32) & 1411.9 
    & 0.9388 \\
    1s CPX & LIBXSMM (FP32) & 1254.8 
    & 0.9387 \\
    1s CPX & LIBXSMM (BF16) & 769.6 
    & 0.9378 \\
    
  \bottomrule
\end{tabular}
\end{table}

\subsection{Scaling Experiments for End-to-End Training}

In this set of experiments, we scale the ATAC-seq training experiments to multiple sockets, longer signal track segment sizes, and larger dataset size.

\subsubsection{\textbf{Multisocket Scaling Experiment}}

\begin{figure}
  \centering
  \includegraphics[scale=0.7]{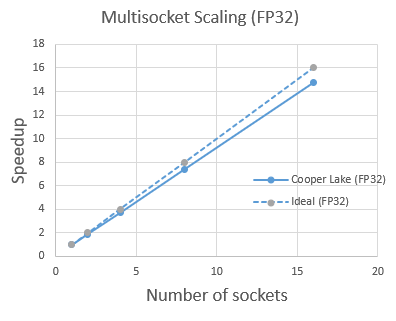}
  \caption{ATAC-seq single-precision (FP32) training time speedup from a single socket to sixteen sockets. Speedup results on sockets of Cooper Lake (CPX).}
  \label{fig:multisocket_scaling_fp32}
\end{figure}

\begin{figure}
  \centering
  \includegraphics[scale=0.7]{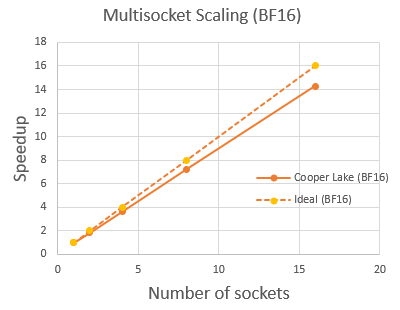}
  \caption{ATAC-seq BFloat16 training time speedup from a single socket to sixteen sockets. Speedup results on sockets of Cooper Lake (CPX).}
  \label{fig:multisocket_scaling_bf16}
\end{figure}

In this experiment, we train the AtacWorks network with the setup described in Section ~\ref{sec:expt-details} on multiple sockets, ranging from one to sixteen sockets. We train the network on \{1,2,4,8,16\} sockets of Cooper Lake CPUs in single-precision and BFloat16 precision. In each run, we train for 25 epochs and collect the training time.
In multi-socket experiments, we reserve one CPU core on each socket for the PyTorch \textit{DataLoader()} worker and one more CPU core on each socket for the message passing interface (MPI) communication, and the other 26 CPU cores for computation. Additionally, we increase the batch size in accordance with the increase in the number of sockets. We keep the batch size as \{54, 52, 104, 208, 416\} for \{1, 2, 4, 8, 16\} socket experiments respectively. Figure~\ref{fig:multisocket_scaling_fp32} shows the speedup for FP32 precision over single-socket training time as we increase the number of sockets. As earlier, in the FP32 experiment, most convolution layers have 15 channels, 15 filters, a filter size of 51, and a dilation of 8. Figure~\ref{fig:multisocket_scaling_bf16} shows the results of the same experiment in BFloat16 precision. In BFloat16 precision, most convolution layers have 16 channels, 16 filters, a filter size of 51, and a dilation of 8. In both figures, we observe that training time scales nearly linearly with an increase in the number of sockets. These scaling results demonstrate that our convolution layer kernels are scalable to multiple sockets. 

\subsubsection{\textbf{Comparison with original AtacWorks running on Nvidia DGX-1 box}}

\begin{figure}
  \centering
  \includegraphics[scale=0.62]{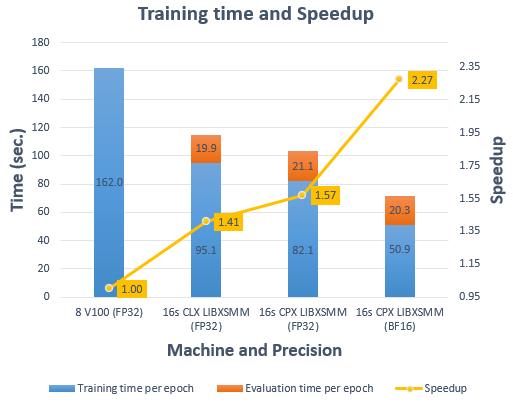}
  \caption{ATAC-seq training time per epoch and speedup results. Timing and speedup results on eight Nvidia V100, sixteen sockets of Cascade Lake (CLX), and sixteen sockets of Cooper Lake (CPX).}
  \label{fig:sixteen socket}
\end{figure}

The AtacWorks paper ~\cite{Lal829481} reports that training the AtacWorks network using the same setup as defined in Section ~\ref{sec:expt-details} for 25 epochs on a DGX-1 box~\cite{nvidia-dgx1} consisting of 8 Nvidia V100 GPUs and a dual socket host CPU takes 2.7 minutes (162 seconds) per epoch. In order to compare with it, we train the AtacWorks network for 25 epochs on 16 sockets of Cascade Lake and Cooper Lake CPUs so as to use nearly the same power envelop. Since it is not clear whether the time reported in ~\cite{Lal829481} includes evaluation time or not, we include the evaluation time for the CPU experiments for comparison. In our training process, we again reserve one CPU core on each socket for the PyTorch \textit{DataLoader()} worker, one more CPU core on each socket for MPI communication, and the other 26 CPU cores on each socket for computation. We use a batch size of 416 to take advantage of 16 CPU sockets. Table~\ref{tab: sixteen socket training} and Figure~\ref{fig:sixteen socket} show training accuracy and time per epoch. Note that our training accuracy for the multi socket runs is nearly the same as that of a single socket run. We can observe that our LIBXSMM library based implementation on sixteen sockets of Cascade Lake achieves up to $1.41\times$ speedup over eight Nvidia V100 GPUs. Sixteen sockets of Cooper Lake (CPX) are $1.57\times$ faster than eight V100 GPUs in training with single-precision (FP32) layers, and they are $2.27\times$ faster with BFloat16 (BF16) layers. Eight sockets of Cooper Lake are also $1.32\times$ faster than eight V100 GPUs with BFloat16 layers. The evaluation is single threaded and doesn't scale, so Figure~\ref{fig:sixteen socket} shows the training and evaluation time separately. Evaluation time is a significant portion of the total time. Therefore, if the time reported for DGX-1 box does not include the evaluation time, our speedups are significantly higher.

\begin{table}[hb]
  \caption{ATAC-seq multi socket training time per epoch (in seconds) on CPUs in comparison with Nvidia V100. Training time includes the evaluation time. The time for 8 V100 GPUs is from ~\cite{Lal829481}. The paper did not report accuracy.}
  \label{tab: sixteen socket training}
  \begin{tabular}{lcccc}
    \toprule
    Device & Precision & Time per & Accuracy & Speedup\\
     &  & epoch (sec.) & (AUROC) & \\
    \midrule
    8 V100 & FP32 & 162.0 & \_ & 1.00x \\
    16s CLX & FP32 & 115.0 & 0.9345 & 1.41x \\
    16s CPX & FP32 & 103.1 & 0.9341 & 1.57x\\
    8s CPX & BF16 & 122.8 & 0.9346 & 1.32x \\
    16s CPX & BF16 & 71.3 & 0.9323 & 2.27x \\
    
  \bottomrule
\end{tabular}
\end{table}

\subsubsection{\textbf{Longer Signal Track Segment Experiment}}

In this experiment, we increase the signal track segment size from 60,000 to 600,000 bases. It increases the input width size of the 1D dilated convolution layer by 10x. Due to the longer signal track segment width, the amount of training data increases as there are fewer nonzero segments. The new training set has 4191 signal track segments, and the new validation set has 101 signal track segments. We finished the training without incurring any out-of-memory error on a dual-socket Cascade Lake. We used a batch size of 52 during the training. The training time per epoch was 977.4 seconds with our optimized implementation.
We were not able to run this experiment on V100 due to GPU's memory constraints.     

\subsubsection{\textbf{Large Dataset Experiment}}

In this experiment, we increase the number of signal track segments to 293242 in the training set and 2520 in the validation set while keeping the signal track segment width as 60000. Thus, the training set in this experiment is approximately $9.16\times$ larger than the one used in previous experiments. We train the AtacWorks neural network for 25 epochs on 16 sockets of Cascade Lake CPUs using our optimized implementation. Our training time per epoch excluding the evaluation time was 872.1 seconds. It is approximately ~$9.16\times$ larger than the training time per epoch with the previous training sets. We were also able to achieve the training accuracy of 0.9390 in terms of the AUROC metric. This demonstrates that our performance scales linearly with the increase in dataset size.

\section{Conclusion}
\label{sec:concl}

Researchers have used 1D convolution layers in multiple fields like audio processing, speech recognition, and genomics. Efficient and generic implementations of the 1D convolution kernels are needed to save time and cost. We believe that future progress in applications like genomics will depend on computing costs. We have shown that code optimizations using the LIBXSMM library's BRGEMM kernel with JIT code generation and cache blocking can increase the efficiency for 1D dilated convolution layers. This approach also helps us generalize the implementations to a wide range of use-cases. Additionally, proper use of Intel\textsuperscript{\textregistered} AVX-512 and AVX-512 BFloat16 instructions can unlock substantial speedups over previous implementations.


\bibliographystyle{ACM-Reference-Format}
\bibliography{sample-base}

\noindent{\small Optimization Notice: Software and workloads used in performance tests may have been optimized for performance only on Intel\textsuperscript{\textregistered} microprocessors. Performance tests, such as SYSmark and MobileMark, are measured using specific computer systems, components, software, operations and functions. Any change to any of those factors may cause the results to vary. You should consult other information and performance tests to assist you in fully evaluating your contemplated purchases, including the performance of that product when combined with other products. For more information go to
http://www.intel.com/performance. Intel\textsuperscript{\textregistered}, Xeon\textsuperscript{\textregistered}, and Intel\textsuperscript{\textregistered} Xeon Phi are trademarks of Intel Corporation in the U.S. and/or other countries.}

\appendix

\end{document}